%% file: id_rep.tex
\documentclass[10pt,twocolumn,letterpaper]{article}

\usepackage{cvpr}                 % Camera-ready (submission: no page numbers)
\usepackage{multicol}
\usepackage{multirow}
\usepackage{makecell}

\input{preamble}
\definecolor{cvprblue}{rgb}{0.21,0.49,0.74}
\usepackage[pagebackref,breaklinks,colorlinks,allcolors=cvprblue]{hyperref}

\def\paperID{51} % *** Enter the Paper ID here
\def\confName{CVPR}
\def\confYear{2026}

\title{Layout-Aware Representation Learning for Open-Set ID Fraud Discovery}

\author{Jinxing Li$^*$, Nicholas Ren, Cathy Chang, Hongkai Pan, Daniel George\\
Persona Identities\\
{\tt\small jinxing.li@withpersona.com}
}

\begin{document}
\maketitle
\input{sec/0_abstract}    
\input{sec/1_intro}
\input{sec/2_related}
\input{sec/3_methodology}
\input{sec/4_results}
\input{sec/5_fraud}
\input{sec/6_conclusions}

\section*{Acknowledgements}
We acknowledge the support from Charles Yeh, Dan Lee, James Chang and Saatvik Billa. 

\section*{Data Statement}
All data used for model training is properly sourced and obtained with appropriate consent for identity verification and fraud prevention purposes. Canada has strict privacy regulations governing the processing of personal data. Accordingly, no data originating from Canadian individuals was used in the training of these models.

{
    \small
    \bibliographystyle{ieeenat_fullname}
    \bibliography{id_rep}
}

% WARNING: do not forget to delete the supplementary pages from your submission 
% \input{sec/X_suppl}

\end{document}

%% file: sec/0_abstract.tex
\begin{abstract}
    Identity-document fraud detection is not a stationary binary classification problem. Adaptive attackers modify templates and fabrication pipelines, making historical fraud labels stale, and successful forgeries recur at scale as coherent campaigns. We therefore study layout-aware representation learning for open-set fraud discovery rather than only closed-set classification. We adapt DINOv3 to the document domain via context-aware SimMIM fine-tuning and supervised metric learning with composite loss that encourages inter-class separability and intra-class compactness. The model is trained with U.S. IDs only. With a lightweight MLP and softmax classifier, the embedding achieves 99.83\% layout classification accuracy on Canadian layouts. Moreover, on a dataset of 20,448 Canadian IDs, embedding-space analysis surfaces 276 adaptive physical-fraud cases, including 222 not surfaced by incumbent detectors. The embedding supports similarity-based expansion from a single confirmed seed to additional related cases not linked by conventional metadata graphs. The layout-aware document embeddings provide a production-aligned basis for discovering novel and campaign-scale fraud under distribution shift.

    \begin{figure}[h]
        \centering
        \includegraphics[width=0.5\textwidth]{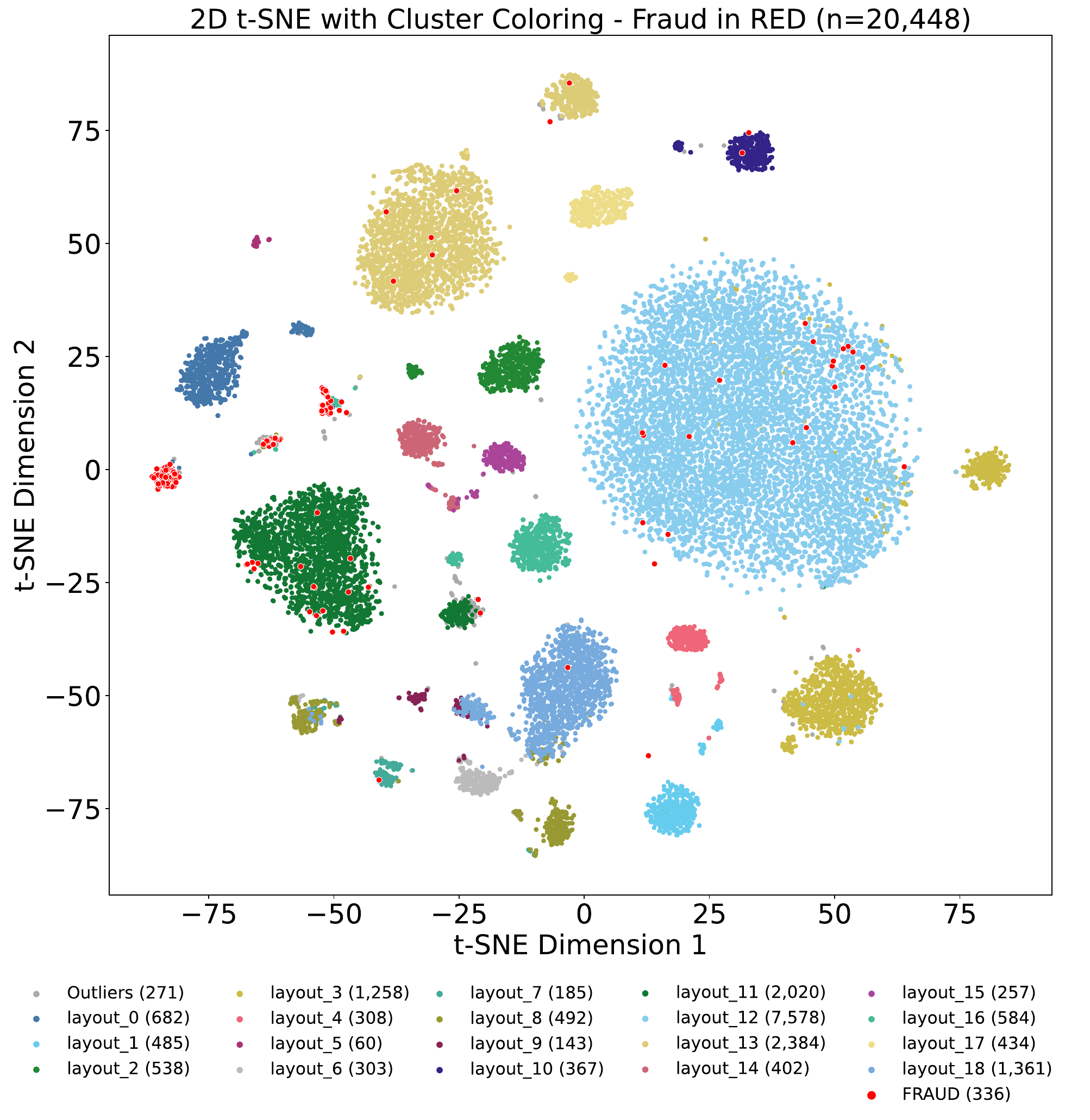}
        \caption{2-D t-SNE visualization of Canadian ID embeddings from real-world data. Although the model was trained only on a U.S. ID dataset, the clustering structure demonstrates strong transferability to previously unseen ID layouts. Three anomalous clusters in the center-left correspond to printed fake passports, illustrating how the embedding space can enable large-scale fraud discovery.}
        \label{fig:fig1}
    \end{figure}
\end{abstract}

%% file: sec/1_intro.tex
\section{Introduction}
\label{sec:intro}

Identity document (IDs) verification is an essential process in online banking, healthcare and business. The rapid advancement of generative AI has made it significantly easier to create and produce fraudulent fake IDs, posing a serious challenge in the Know Your Customer (KYC) process. Traditional machine learning treatments of fraud are framed as supervised classification on static datasets. That framing is often mismatched to real-world production identity fraud. In production identity verification, fraud spans at least two operational regimes. The first is \textbf{opportunistic (non-adaptive)} fraud, where submissions are low-effort or visually obvious and are often well handled by supervised classifiers or rules trained on historical examples. The second is \textbf{adaptive (adversarial)} fraud, where attackers adapt to defenses, modify document templates and fabrication pipelines, and quickly exploit newly successful attack patterns at scale. This second regime is more challenging because it is non-stationary (future fraud traffic is not fully represented by historical labeled fraud), weakly labeled, and often campaign-structured. Our paper focuses on this adaptive regime. In this setting, a useful model must do more than scoring isolated documents; it must surface previously unseen fraud families, expand from a small number of confirmed seeds, and enable cluster-level triage by investigators (e.g., \cite{lunghi2023adversarial}). 

One challenge of fraudulent ID layout detection is the vast number of possible layouts which are ever-growing and highly imbalanced. Many countries maintain a large number of supported ID layouts issued by the national and subdivisions. For instance, the United States alone has at least 886 supported ID layouts as of 2026. Existing datasets cover only a small fraction of these variants, and newly issued layouts are not labeled in advance, requiring models to generalize effectively to unseen layouts.

Secondly, in real-world scenarios, ID documents are captured using mobile phone cameras under various conditions, including different illumination, orientation, noise and backgrounds. Real-world ID images may exhibit a number of adverse scenarios that impact image quality further complicating the ID layout and fraud recognition task. 

Finally, labeled data is costly to obtain due to privacy obligations and the diversity of document designs. Maintaining comprehensive, up-to-date annotations becomes increasingly impractical. These challenges motivate approaches that can leverage unlabeled data, learn discriminative representations, and generalize beyond seen document layouts.

ID layouts exhibit high intra-class variance and low inter-class variance. This requires the learning of fine-grained, discriminative features that capture the invariant structural layout of the document. We propose a ID layout training framework that leverages self-supervised pre-training and supervised metric learning. Our contributions are summarized below:
\begin{itemize}
\item We introduce a layout-aware document embedding model that supports similarity search, clustering, and centroid-based anomaly scoring for fraud analysis.
\item We show strong cross-domain transfer from U.S. training data to Canadian IDs. By adding a lightweighted MLP to the embedding, the model achieves 99.83\% classification accuracy on Canadian IDs. 
\item We demonstrate concrete anti-fraud value in production traffic, including identifying anomalous clusters and similarity-based expansion from a single confirmed fraud seed that surfaces numerous sophisticated fraud uncaught by incumbent supervised models.
\end{itemize}

%% file: sec/2_related.tex
\section{Related Work}
\label{sec:related_work}

%-------------------------------------------------------------------------
\subsection{Fraud Detection Under Distribution Shift and Campaign Structure}
Fraud detection systems operate in a hostile environment where attackers actively adapt to deployed defenses \cite{lunghi2023adversarial}. Fraud detection in production differs substantially from traditional calssification, because fraud strategies evolve over time, while labels are delayed and incomplete, and only a subset of alerts receives prompt expert review. Hence, models trained on historical fraud examples quickly become stale under distribution shift. For instance, credit card system focus on  concept drift adaption with delayed supervision information rather than  static classification on a fixed dataset \cite{Pozzolo2015}.

Classic classification is effective for closed-set, as conclusions drawn from small or simplified benchmarks may be transferrable to more realistic large-scale settings. Open-set recognition (OSR), however, addresses the scenario in which test-time inputs may not belong to any class observed during training \cite{bendale2016towards}. Handling test-time distribution shift is a central requirement for safe deployment (\cite{Mahdavi_2021, hendrycks2022scalingoutofdistributiondetectionrealworld, yang2022openoodbenchmarkinggeneralizedoutofdistribution, wang2024dissectingoutofdistributiondetectionopenset}). In the ID fraud domain, new layouts, fabrication pipelines, and manipulation strategies continually emerge, hence, effective systems must be capable of surfacing previously unseen fraud families.

In addition, frauds in the real world are structured and collective rather than independent across instances. In e-commerce and payment systems, graph-based and collective methods are developed to exploit inter-instance dependency, coordinated behavior, and group structure to identify fraud that may appear benign when each case is scored in isolation \cite{hooi2016fraudar, cao2018collective, yu2023}. Previous forensic-document research describes fraudulent identity documents as having a serial character, where multiple documents share a common source and can be detected as a series for intelligence and investigative purposes, including in cross-border settings \cite{MOULIN2022, MOULIN2024}.

\subsection{Self-Supervised Learning with Masked Image Modeling}
\paragraph{Self-supervised learning (SSL)}  has emerged as a powerful paradigm to learn transferable visual representations without human annotations. Early SSL approaches focused on handcrafted pretext tasks, such as predicting image rotations \cite{gidaris2018unsupervisedrepresentationlearningpredicting}, relative patch positions \cite{noroozi2017unsupervisedlearningvisualrepresentations}, image inpainting and context prediction \cite{pathak2016contextencodersfeaturelearning}, and split-brain autoencoding objectives \cite{zhang2017splitbrainautoencodersunsupervisedlearning}. 

Contrastive learning is a popular SSL framework. Methods such as SimCLR learn representations by maximizing agreement between differently augmented views of the same image while contrasting them against other samples in the batch \cite{chen2020simpleframeworkcontrastivelearning}. Momentum Contrast (MoCo) further improves scalability by introducing a momentum-updated encoder and a dynamic queue of negative samples, reducing the need for extremely large batch sizes \cite{he2020momentumcontrastunsupervisedvisual}.

Another important class of SSL methods is generative or reconstruction-based learning. Masked image modeling approaches, such as Masked Autoencoders (MAE), reconstruct missing image regions from partially observed inputs and have demonstrated strong scalability and transfer performance for Vision Transformers \cite{he2021maskedautoencodersscalablevision}. In contrast, SimMIM employs a standard Vision Transformer (ViT) that processes the full image—including masked tokens—and predicts the raw pixel values of masked patches using a lightweight reconstruction head \cite{Xie_2022_CVPR}.

\paragraph{DINO} Beyond contrastive learning with explicit negative samples, the Bootstrap Your Own Latent (BYOL) \cite{grill2020bootstraplatentnewapproach} model introduces a non-contrastive framework based on self-distillation. Building upon the self-distillation paradigm, DINO (self-DIstillation with NO labels) \cite{caron2021dino} extends BYOL by formulating the learning objective as a distribution matching problem. In DINO, a student network is trained to match the output distribution of a teacher network whose parameters are updated via an exponential moving average of the student weights. The model processes multiple augmented views of the same image and optimizes their alignment using a cross-entropy loss. The DINO yields semantically meaningful attention maps that often correspond to object-centric regions, enabling emergent object localization without supervision.

DINOv2 \cite{oquab2024dinov2} scaled to larger model sizes and significantly larger, curated and diversified training datasets. DINOv3 \cite{siméoni2025dinov3} further scaled the model sizes and training datasets, improved training stability, scalability, and representation quality. It has demonstrated strong performance across a wide range of downstream tasks, including image classification, object detection, and segmentation, often without task-specific fine-tuning. It has been widely adopted as a general-purpose visual backbone and foundation model for downstream tasks, e.g., medical vision tasks \cite{liu2026doesdinov3setnew}, robotics \cite{egbe2025dinov3diffusionpolicyselfsupervisedlarge} and object detection \cite{huang2025realtimeobjectdetectionmeets}.

\subsection{Deep Metric Learning for Fine-Grained Classification}

\paragraph{Deep metric learning} learns an embedding space where semantically similar samples are close and dissimilar ones are separated by large margins. It is particularly effective for fine-grained classification tasks where class distinctions are subtle, such as face recognition. Early approaches rely on pairwise or triplet objectives such as contrastive loss \cite{hadsell2006dimensionality} and triplet loss \cite{schroff2015facenet} that enforce similarity constraints between pairs or triplets of samples. However, these methods often converge slowly and require careful sample mining strategies \cite{wu2017sampling}. Recent work instead adopts classification-based metric learning losses that incorporate geometric constraints directly into the softmax objective, such as SphereFace \cite{liu2017sphereface}, CosFace \cite{wang2018cosface}, and ArcFace \cite{deng2019arcface}.

\paragraph{ArcFace Loss} introduces an additive angular margin to the softmax loss to enhance inter-class separability in the embedding space \cite{deng2019arcface}. Both feature vectors and classifier weights are normalized so that classification depends solely on angular distances on a hypersphere. Let $W_j$ denote the normalized weight vector for class $j$, and let $\theta_j$ be the angle between $z_i$ and $W_j$. The ArcFace loss is defined as
\begin{equation}
\mathcal{L}_{\text{ArcFace}} = -\frac{1}{N} \sum_{i=1}^{N} \log \frac{ e^{s\cos(\theta_{y_i}+m)} }{ e^{s\cos(\theta_{y_i}+m)} + \sum_{j \neq y_i} e^{s\cos(\theta_j)} }
\end{equation}
where $s$ is a feature scaling factor and $m$ is the additive angular margin. By explicitly increasing the angular distance between classes, ArcFace produces highly discriminative embeddings and has become a standard loss for large-scale face recognition.

\paragraph{Supervised Contrastive Learning(SupCon)} extends contrastive learning to the supervised setting by leveraging label information to define positive and negative pairs within a batch \cite{khosla2021supervisedcontrastivelearning} . For an anchor embedding $z_i$, the set of positives $P(i)$ includes all samples in the batch that share the same label. The SupCon loss is defined as
\begin{equation}
\mathcal{L}_{\text{SupCon}} = \sum_{i=1}^{N} \frac{-1}{|P(i)|} \sum_{p \in P(i)} \log \frac{ \exp(\mathbf{z}_i \cdot \mathbf{z}_p / \tau) }{ \sum_{a \neq i} \exp(\mathbf{z}_i \cdot \mathbf{z}_a / \tau) }
\end{equation}
where $|P(i)|$ is the number of positive samples in the batch, and $\tau$ is a temperature parameter controlling the concentration of the distribution. By pulling together embeddings from the same class and pushing apart embeddings from different classes, SupCon effectively utilizes multiple positive samples per anchor, leading to more stable and discriminative representations.

\paragraph{Center Loss} While margin-based softmax losses improve inter-class separation, they do not explicitly enforce intra-class compactness. Center Loss \cite{wen2016discriminativefeaturelearningapproach} addresses this limitation by learning a center $c_k$ for each class $k$ and minimizing the distance between features and their corresponding class centers:
\begin{equation}
\mathcal{L}_{\text{center}} = \frac{1}{2} \sum_{i=1}^{N} \left\| \mathbf{z}_i - \mathbf{c}_{y_i} \right\|_2^2
\end{equation}
During training, class centers are updated iteratively using mini-batch statistics. This formulation encourages features from the same class to form tight clusters in the embedding space.

%% file: sec/3_methodology.tex
\section{Methodology}

\begin{figure*}[t]
    \centering
    \includegraphics[width=\textwidth]{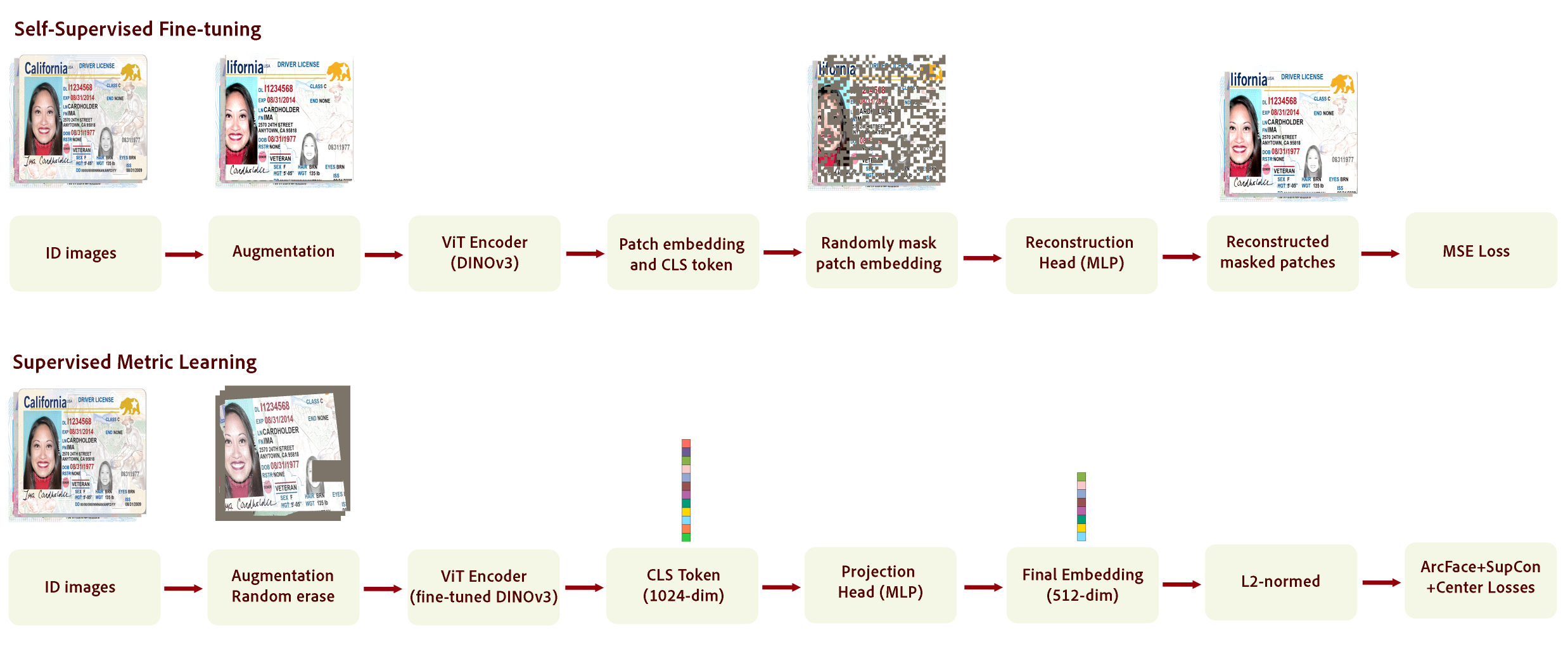}
    \caption{Training framework for layout-aware ID representation learning. Top: self-supervised fine-tuning adapts the DINOv3 Vision Transformer to the document domain using SimMIM masked reconstruction. Bottom: Supervised metric learning maps ID images to 512-dimensional embeddings via a projection head, and the network is trained with a composite loss (ArcFace, supervised contrastive, and center loss) to encourage inter-class separability and intra-class compactness.}
    \label{fig:fig2}
\end{figure*}

We first adapt the general-purpose DINOv3 model to the document domain via context-aware self-supervised pre-training. Then, we further refine the learned representations via supervised metric learning with a composite loss function, yielding a highly discriminative embedding space suitable for downstream tasks such as classification, retrieval, clustering, and new set recognition. The training framework is shown in Figure \ref{fig:fig2}.

\subsection{Self-Supervised Fine-tuning}
Since the web-scale training of DINOv3 provides a robust, heavily generalizable prior for visual features, we adopt the pretrained DINOv3 weights as the initialization of our model's encoder. We use U.S. ID images for the training.
  
\paragraph{Image Augmentation} To improve robustness to real-world capture conditions and prevent overfitting, we apply strong data augmentation before the fine-tuning. The augmentation simulates imperfect framing, rotation, skew, and camera viewpoint changes commonly observed in mobile and handheld document capture. It preserves global layout semantics while introduces substantial appearance and geometric variability. We first resize each input image to 640x640. This is followed by mild color jittering, where brightness, contrast, and saturation are each independently perturbed within ±10\%, and hue is perturbed within ±0.02. The augmented images are then normalized by the channel-wise mean and standard deviation of ImageNet statistics \cite{deng2009imagenet}.

\paragraph{Model Architecture}  The DINOv3 encoder produces the patch embedding and the 1024-dimension global CLS token embedding for augmented ID images. We randomly mask 60\% of the patch embeddings, and use the remaining embeddings to reconstruct the original image. This design allows the self-attention mechanism to construct contextualized representations for all patches, informed by the global document layout. The reconstruction head is a Multi-Layer Perceptron (MLP) that predicts their original pixel values. 
  
\paragraph{Fine-tuning Strategy} To stabilize adaptation and reduce overfitting, we use progressive unfreezing. Initially, the encoder is frozen and only the reconstruction head is trained with AdamW. Starting at epoch 1, one transformer block per epoch (block 23 → 12) is unfrozen from top to bottom and added to the optimizer as a new parameter group. Training runs for 30 epochs, and the best model is selected by validation loss.

\subsection{Supervised Metric Learning}
We implement supervised learning to learn a structured embedding space that is highly discriminative for seen layouts, meanwhile generalizable to unseen layouts. We use 15k labeled IDs from almost all U.S. states. We split these layouts into disjoint training, validation, and test sets by states of origin. A batch size of 64 is used for training.

\paragraph{Data Augmentation} We first apply Geometric Augmentations, including random resized cropping (covering 60--100\% of the image area), affine transformations including rotation (up to $20^\circ$), translation (up to 15\%), shear, and probabilistic perspective distortion. We then apply Photometric Augmentations, including color jittering to brightness, contrast, saturation, and hue, followed by mild Gaussian blur to simulate variation in illumination, printing, and camera sensors. Finally, images are normalized using ImageNet statistics, and are randomly masked to discourage reliance on localized visual cues and encourage learning of global layout structure.

\paragraph{Model Architecture and Composite Loss Function}  We used the fine-tuned DINOv3 encoder (discard the reconstruction head) as the base. The encoder outputs a 1024-dimensional CLS token embedding, which is passed through a projection head implemented as a light-weighted MLP with batch normalization and dropout. The projection head reduces the dimensionality to a 512-dimensional embedding, which serves as the final representation for downstreaming tasks. To jointly encourage inter-class separability and intra-class compactness, we employ a composite loss function:
\begin{equation}
\mathcal{L}_{\text{total}} = \lambda_A \mathcal{L}_{\text{ArcFace}} + \lambda_S \mathcal{L}_{\text{SupCon}} + \lambda_C \mathcal{L}_{\text{Center}}
\end{equation}
where $\lambda_A$, $\lambda_S$, and $\lambda_C$ are the weights for the ArcFace loss, Supervised Contrastive loss, and Center loss, respectively. We will discuss the choice of these weights in the next section.
\paragraph{Training Strategy}  Our training strategy follows the widely adopted gradual unfreezing paradigm in transfer learning, where task-specific heads are first optimized while keeping the pretrained backbone frozen, followed by progressively unfreezing deeper layers. This staged approach has been shown to stabilize optimization and mitigate catastrophic forgetting when adapting pretrained models to new tasks \cite{howard2018universal, montone2017gradual}.
\begin{itemize}
\item Warm-up: The encoder backbone is frozen. Only the Projection and ArcFace heads are trained ($LR=5 \times 10^{-4}$). This establishes an initial cluster structure.
\item Partial Unfreezing: The last 8 blocks of the ViT backbone are unfrozen progressively ($LR=1 \times 10^{-5}$).
\item Full Fine-tuning: All layers are unfrozen with a reduced learning rate ($10\%$ of Partial Unfreezing) to refine the global feature extractors.
\end{itemize}

%% file: sec/4_results.tex
\section{Experiment Results}

We constructed a test set using an unseen dataset to evaluate its generalization performance. Specifically, we use a Canadian ID dataset collected from a real-world scenario.
\subsection{ID clustering and annotation}
Manual ID labeling is labor-intensive, time-consuming, and difficult to scale as new document layouts are continuously introduced. Since ID embeddings typically exhibit high intra-class variance and low inter-class variance, we leverage clustering to group a large collection of IDs in the embedding space, enabling bulk pseudo-labeling with minimal manual intervention.

We apply k-means clustering to the learned ID embeddings and visualize the resulting distribution using 2D t-SNE (like that shown in Figure \ref{fig:fig1} but are different datasets). Major layouts naturally form well-separated clusters, indicating that the embeddings capture discriminative layout-level features. Clustering provides a strong initialization for large-scale annotation, significantly reducing manual effort to reviewing and correcting misclustered samples.

\subsection{Model performance on test set}
The supervised training minimizes intra-class distance while maximizing inter-class distance, hence, are these two metrics are examined in the ablation study. To evaluate the clustering quality, we employ the Silhouette score defined as
\begin{equation}
s(i)=\frac{b(i)-a(i)}{max(a(i),b(i))}
\end{equation}
where $a(i)$ is the average distance between sample $i$ and all other samples in the same class (intra-class distance), and  $b(i)$ is the minimum average distance between sample $i$ and all samples in nearest neighbouring class (the next best alternative). We further employ the Davies--Bouldin index (DBI) to quantify the clustering validity, which is defined as 
\begin{equation}
\mathrm{DBI} = \frac{1}{K} \sum_{i=1}^{K} \max_{j \neq i}
\left( \frac{S_i + S_j}{M_{ij}} \right),
\end{equation}
where $K$ is the number of clusters, $S_i$ denotes the within-class scatter of cluster $i$, and $M_{ij}$ is the distance between the centroids of class $i$ and $j$.

Table \ref{tab:table1} exhibits the clustering metrics on the Canadian ID test set. Comparing the original and the fine-tuned domain-specific DINOv3 models, the fine-tuned model achieves a higher Silhouette score (0.132 vs. 0.123) and a lower DBI (2.501 vs. 2.718), indicating superior cluster compactness and separation on unseen ID layouts. Given this improved embedding quality, we adopt the fine-tuned model for subsequent supervised discriminative training.

On supervised learning, we evaluate four configurations with different loss formulations: (1) ArcFace loss only, (2) SupCon loss only, (3) ArcFace + SupCon, and (4) ArcFace + SupCon + Center loss. The corresponding weighting coefficients ($\lambda_A$, $\lambda_S$, and $\lambda_C$) are listed in Table \ref{tab:table1}. We use an equal weighting for ArcFace and SupCon losses since both are log-softmax on cosine similarities. For the center loss, we adopt a $\lambda_C = 0.003$, which is originally used by Wen et al.~\cite{wen2016discriminativefeaturelearningapproach}. Experiments show that the combination of all three losses yields the highest Silhouette score and the lowest DBI, demonstrating improved intra-class compactness and inter-class separability. This configuration constitutes our best-performing model.

To assess whether self-supervised fine-tuning is necessary, we conduct an ablation study using the original DINOv3-Large model as the base model. We then apply supervised fine-tuning with the same loss configurations, where the best result is again obtained using the combination of all three losses (for brevity, only this best variant is shown in Table \ref{tab:table1}). Although this model achieves a marginally smaller intra-class distance (0.444 vs. 0.445), its Silhouette score and DBI are inferior to those of the domain-fine-tuned counterpart. This suggests that self-supervised fine-tuning on real-world ID images enables the model to capture richer layout-specific contextual and spatial representations, providing a stronger initialization than the original DINOv3 model.

The ID datasets used for training and the ablation studies described above are proprietary. To facilitate evaluation on a public benchmark, we further assess our approach on the 2025 DeepID Challenge dataset \cite{Korshunov_2025_ICCV}, which contains approximately 4,000 FantasyID images with diverse layouts captured using different devices \cite{vidit2025fantasyiddatasetdetecting}. We evaluate our model on the bona fide subset, and the results are summarized in Table \ref{tab:table2}.

Interestingly, the fine-tuned domain-specific DINOv3 achieves the best Silhouette (0.693) and DBI (0.431) on this benchmark. However, we emphasize that the FantasyID dataset does not fully reflect real-world ID verification conditions. This difference is reflected in the clustering statistics: the Silhouette score on FantasyID (0.693) is substantially higher than that on the real-world ID dataset (0.311), indicating a much cleaner and less noisy distribution. Therefore, while FantasyID serves as a useful public benchmark, its relative homogeneity may not capture the full complexity and distributional challenges encountered in practical deployment settings.

\begin{table*}[t]
    \centering
    \caption{The mean intra-class distance, mean inter-class distance, Silhouette score, and DBI of the test set Canadian IDs, calculated for the DINOv3, the fine-tuned DINOv3, and supervised learning models.}
    \label{tab:table1}
    \renewcommand{\arraystretch}{1.3}
    \begin{tabular}{lcccc}
        \toprule
        \textbf{Model}
            & \textbf{Intra-class} $\downarrow$ 
            & \textbf{Inter-class} $\uparrow$ 
            & \textbf{Silhouette} $\uparrow$ 
            & \textbf{DBI} $\downarrow$ \\
        \midrule

        \textbf{DINOv3}         & 0.446 & 0.655 & 0.123 & 2.718 \\

        \quad $\lambda_A{=}1, \lambda_S{=}1, \lambda_C{=}0.003 $
            & 0.444 & 1.163 & 0.303 & 1.403 \\
        \midrule
        \textbf{Fine-tuned DINOv3}    & \textbf{0.377} & 0.612 & 0.132 & 2.501 \\

        \quad$\lambda_A{=}1, \lambda_S{=}0, \lambda_C{=}0$
        & 0.493 & 1.124 & 0.289 & 1.354 \\

        \quad$\lambda_A{=}0, \lambda_S{=}1, \lambda_C{=}0$
        & 0.669 & 1.179 & 0.180 & 2.033 \\

        \quad$\lambda_A{=}1, \lambda_S{=}1, \lambda_C{=}0$
        & 0.502 & \textbf{1.219} & 0.282 & 1.355 \\

        \quad$\lambda_A{=}1, \lambda_S{=}1, \lambda_C{=}0.003$
        & 0.445 & 1.168 & \textbf{0.311} & \textbf{1.265} \\
        \bottomrule
    \end{tabular}
\end{table*}

\begin{table*}[t]
    \centering
    \caption{The intra-class distance, inter-class distance, Silhouette score, and DBI of the FantasyID dataset.}
    \label{tab:table2}
    \renewcommand{\arraystretch}{1.3}
    \begin{tabular}{lcccc}
        \toprule
        \textbf{Model}
            & \textbf{Intra-class} $\downarrow$ 
            & \textbf{Inter-class} $\uparrow$ 
            & \textbf{Silhouette} $\uparrow$ 
            & \textbf{DBI} $\downarrow$ \\
        \midrule

        \textbf{DINOv3}         & 0.349 & 0.676 & 0.386 & 1.062 \\

        \quad $\lambda_A{=}1, \lambda_S{=}1, \lambda_C{=}0.003$
            & 0.286 & \textbf{1.091} & \textbf{0.693} & \textbf{0.431} \\
        \midrule
        \textbf{Fine-tuned DINOv3}    & 0.290 & 0.643 & 0.410 & 0.988 \\

        \quad$\lambda_A{=}1, \lambda_S{=}0, \lambda_C{=}0$
            & 0.312 & 1.120 & 0.669 & 0.490 \\

        \quad$\lambda_A{=}0, \lambda_S{=}1, \lambda_C{=}0$
            & 0.309 & 0.984 & 0.588 & 0.690 \\

        \quad$\lambda_A{=}1, \lambda_S{=}1, \lambda_C{=}0$
            & 0.286 & 1.031 & 0.667 & 0.537 \\

        \quad$\lambda_A{=}1, \lambda_S{=}1, \lambda_C{=}0.003$
            & \textbf{0.269}  & 1.063 & 0.688 & 0.473 \\
        \bottomrule
    \end{tabular}
\end{table*}

\subsection{ID layout classification}
To evaluate the discriminative power of the learned embeddings for downstream tasks, we perform ID layout classification for the Canadian ID dataset that contains 19 primary layouts (rarely-seen layouts are excluded for reliable evaluation). We append a lightweight multilayer perceptron (MLP) head to the embedding, consisting of a 256-unit hidden layer, a ReLU activation and a dropout layer with a rate of 0.2. We adopt an 80/20 split within each layout category to ensure balanced supervision across classes. The model is trained using cross-entropy loss. On the held-out test set, the classifier achieves 99.83\% overall accuracy, demonstrating strong generalization across diverse layouts. In contrast, the original DINOv3 model achieves 99.32\% overall accuracy, indicating that the supervised fine-tuning significantly improves the model's ability to generalize to unseen layouts.

Although the current experiment focuses on a moderate-scale layout taxonomy, real-world global ID systems may involve hundreds or thousands of distinct layouts. Scaling layout classification to such a large and dynamically evolving label space introduces additional challenges, including long-tail distribution effects and fine-grained inter-layout similarity. We leave comprehensive evaluation on large-scale global ID datasets as future work.

%% file: sec/5_fraud.tex
\section{Open-Set Fraud Discovery in Production Traffic}

Adaptive fraud, in which attackers modify artifacts to evade deployed models, is the primary target of this work. This setting differs fundamentally from fraud detection on closed datasets such as FantasyID, where fraud categories are predefined and stable. Hence, we do not measure success by overall recall across all confirmed fraud cases, since that metric conflates easy and difficult regimes and is further confounded by incomplete production labels. Instead, we focus on incremental discovery: confirmed fraud surfaced by the embedding method beyond the incumbent stack, especially within the adaptive physical-fraud regime.

Fraud labels in our experiments are obtained through a multi-source ground-truth pipeline that combines programmatic signals with expert human review. The programmatic pipeline maintains a running set of fraudulent verifications identified via existing rule-based fraud signals, escalation workflows, and cases that trigger multiple downstream fraud checks simultaneously. Expert visual inspection is critical for validating cases that automated signals may not fully capture. Commonly inspected indicators include layout inconsistencies, saturation anomalies, color artifacts suggestive of digital replicas, evidence of tampering or image compositing, and portrait irregularities, image manipulation, etc. Fraud labels are most confidently confirmed when programmatic signals and human inspection concur, whereas benchmarking of new detection methods is conducted against the automated programmatic labels.

\subsection{Similarity-based fraud discovery}
The learned embedding space enables an effective similarity-driven fraud discovery mechanism. In practice, fraudulent IDs generated from a common template or fabrication pipeline often share similar features, such as suturation, fontsize, fontface, spacing tweaks, texture, etc. Their embeddings deviate from the centroid of their layout, but they exhibit strong mutual similarity.

Empirically, this embedding-based graph node substantially improves fraud discovery. Starting from a single confirmed fraudulent ID (a seed), the similarity edges enable the identification of related fraudulent documents that are not connected through conventional graph signals (e.g., shared metadata, device fingerprints, or user attributes). This demonstrates that visual embedding similarity captures complementary fraud patterns that are otherwise difficult to detect.

\subsection{Cluster-level fraud discovery}
In real-world fraud, successful fabrication pipelines are typically scaled, producing many visually consistent submissions. This induces redundant evidence: investigators can validate a suspected fraud campaign by inspecting a small number of representative samples from a cluster, rather than adjudicating each document independently. As a result, we evaluate our method not only by per-document scoring performance, but also by label efficiency—the number of fraudulent submissions identified per unit of human review—where a single cluster-level decision can cover tens to hundreds of related attempts. This human-in-the-loop framing allows the embedding model to prioritize high-recall discovery of coherent groups, while maintaining manageable review cost.

The global structure of the embedding space can reveal previously unknown fraud patterns. To study this structure, we cluster 20,448 real-world Canadian ID embeddings. We first project the embeddings into 2D space using t-SNE for visualization, and then perform k-means clustering with a suitable number of clusters. The resulting clusters are iteratively refined over multiple auditing rounds through splitting, merging, outlier removal, and percentile-based trimming to improve intra-cluster cohesion. Such outlier handling is especially important in real-world settings, where noisy samples are common.

Figure \ref{fig:fig1} shows a two-dimensional t-SNE projection of the embeddings colored by their cluster assignments. Fraudulent IDs (marked in red) are distributed across multiple layouts.

A particularly notable observation is the emergence of three dense clusters located far from primary legitimate layout centroids. Manual investigation reveals that these samples correspond to printed passport reproductions that were submitted as ID documents, and different clusters represent different passport layouts.

Remarkably, these three anomalous clusters surfaced a total of 276 confirmed cases of fraud. Incumbent supervised models and conventional metadata-graph signals only captured 54 of them. This finding highlights the effectiveness of embedding-space clustering for uncovering large-scale fraud operations and systematic fabrication sources, especially when no prior template labels are available.

\subsection{Borderline and High z-Score Cases}
We investigated samples in a cluster but are farthest from the centroids (having largest z-score where $z=(x-\mu)/\sigma$). We found many of them belong to the following situations. (1) Blurred, out of focus, snapshot during moving, and low resolution due to dim lighting condition, making it difficult or impossible to read ID information; (2) replay from screen, showing clear Moiré pattern; (3) physically printed IDs, paper printed IDs and deepfakes. The z-score alone can assist us detect unreal IDs.
Even without pairwise similarity analysis, the z-score itself provides a useful anomaly signal for identifying suspicious or low-quality submissions. Combined with similarity-based graph propagation and clustering analysis, the embedding space offers a unified and scalable framework for real-world fraud discovery.

%% file: sec/6_conclusions.tex
\section{Conclusions and Discussions}
This work presents an ID layout representation model by adapting a DINOv3 Vision Transformer foundation model to the ID domain through context-aware self-supervised fine-tuning, and supervised discriminative learning with a composite loss formulation. The embedding space is well-suited for downstream tasks including classification, retrieval, clustering, and open-set recognition. Experiments show that the model can classify real world ID layouts with high accuracy.

Fraudulent IDs originating from a common fabrication source exhibit consistent visual artifacts, and form coherent structures in the learned embedding space. Our model makes authentic layouts compact while allowing coherent fraud sources to emerge as off-manifold clusters or high-similarity subgraphs, and proves effective in discovering previously unseen fraud families and scaled fraud campaigns in an adversarial and non-stationary fraud settings. By incorporating embedding similarity into a graph-based detection framework, we can propagate signals from a confirmed fraudulent ID to a broader cluster of related fraud cases. 

The proposed embedding framework provides a production-aligned foundation for human-in-the-loop fraud discovery. While this work focuses on identity documents, the approach naturally generalizes to other semi-structured documents, such as bank statements, pay stubs, tax forms, and certificates, where layout consistency and subtle structural deviations are critical for authenticity verification.